\documentclass{article}

\usepackage{arxiv}

\usepackage[utf8]{inputenc} % allow utf-8 input
\usepackage[T1]{fontenc}    % use 8-bit T1 fonts
\usepackage{hyperref}       % hyperlinks
\usepackage{url}            % simple URL typesetting
\usepackage{booktabs}       % professional-quality tables
\usepackage{amsfonts}       % blackboard math symbols
\usepackage{nicefrac}       % compact symbols for 1/2, etc.
\usepackage{microtype}      % microtypography
\usepackage{lipsum}
\usepackage{svg}
\usepackage{graphicx}
\graphicspath{ {./images/} }
\usepackage{amsmath} 
\usepackage{pdflscape}
\usepackage{xcolor}
\usepackage[normalem]{ulem}
\usepackage{comment}
\usepackage{longtable}
\usepackage{tabularx}
\usepackage{caption} % for more complex captioning
\usepackage{subcaption}
\usepackage{longtable}
\usepackage{booktabs}
\usepackage[utf8]{inputenc}
\usepackage{changepage}

\usepackage{array}
\usepackage{float}
\newlength{\extralength}
\newlength{\fulllength}
\newcolumntype{C}{>{\centering\arraybackslash}X}

\title{What is YOLOv6? A Deep Insight into the Object Detection Model}

\author{Athulya Sundaresan Geetha\\[1ex]
\begin{minipage}[t]{0.90\textwidth}
\centering
\scriptsize Department of Computer Science, Huddersfield University, Queensgate, Huddersfield HD1 3DH, UK; \\
\textsuperscript{*}Correspondence: U2282847@unimail.hud.ac.uk;
\end{minipage}}

\begin{document}

\maketitle
% Abstract (Do not insert blank lines, i.e. \\) 
\begin{abstract}This work explores the YOLOv6 object detection model in depth, concentrating on its design framework, optimization techniques, and detection capabilities. YOLOv6's core elements consist of the EfficientRep Backbone for robust feature extraction and the Rep-PAN Neck for seamless feature aggregation, ensuring high-performance object detection. Evaluated on the COCO dataset, YOLOv6-N achieves 37.5\% AP at 1187 FPS on an NVIDIA Tesla T4 GPU. YOLOv6-S reaches 45.0\% AP at 484 FPS, outperforming models like PPYOLOE-S, YOLOv5-S, YOLOX-S, and YOLOv8-S in the same class. Moreover, YOLOv6-M and YOLOv6-L also show better accuracy (50.0\% and 52.8\%) while maintaining comparable inference speeds to other detectors. With an upgraded backbone and neck structure, YOLOv6-L6 delivers cutting-edge accuracy in real-time.
\end{abstract}

% Keywords

\section{Introduction}
Computer vision is a fast-moving field that provides machines \cite{RN100,hussain2023custom} with the ability to process and comprehend visual signals. Object detection is a crucial element of this field, involving the accurate localization and identification of objects within images or video streams \cite{RN102}.

Features were extracted from images both horizontally and vertically using the Histogram of Oriented Gradients (HOG) and Vertical Histogram of Oriented Gradients (VHOG) techniques \cite{RN1}. Support Vector Machine and Extreme Learning Machine were then used for classification. The SVM method relied on visualizing features by localizing images and features during dynamic motion in spatiotemporal areas for classification but faced challenges in accurately detecting movements \cite{RN2, RN3, RN4,hussain2022feature}.

The inefficiencies of two-stage approaches, such as their extended runtime, dependence on advanced architectures, and need for constant manual intervention, led to the rise of CNN models for classification and detection. When used together, Faster R-CNN and GoogleNet achieved better results in object detection than the previous models \cite{RN5}. PoseNet, MaskR-CNN, and MobileNet were used in another study to refine image segmentation and boost performance. Other models, such as Region-based Convolutional Network method (R-CNN) \cite{RN10}, Google-Net \cite{RN7}, Fast-RCNN \cite{RN11}, AlexNet \cite{RN6}, ResNet \cite{RN8}, Faster-RCNN \cite{RN12}, and VGG-Net \cite{RN9}, were used to improve object identification further \cite{RN13}.

Even with performance improvements, challenges such as efficient data management, real-time execution, and more advanced architectures persist. YOLO (You Only Look Once) models address these limitations, overcoming the drawbacks of two-stage detection methods. YOLOv1, based on DarkNet, includes 24 convolutional layers, and Fast YOLO includes 9 layers \cite{RN14}. YOLOv2 is a refined version built on DarkNet-19, providing batch standardization, box bounding optimization, in-depth class identification, intricate feature extraction ~\cite{hussain2022gradient}, and size-oriented dimension \cite{RN15}. YOLOv3, based on the DarkNet-53 architecture, employs residual networks for efficient feature extraction and binary cross-entropy for training optimization \cite{RN16}. YOLOv4, built on YOLOv3’s predictive head, features CSPDarkNet53 and integrates PANet with SPP as the backbone and neck for enhanced performance, respectively \cite{RN17}.

The high-performance speed of YOLOv5 is achieved through the combination of the enhanced CSP-PAN neck and the SPPF head \cite{RN18}. Not only does YOLOv6 offer enhanced performance speed, but it also reduces the complexity of computational processes \cite{RN19}. With the addition of an auxiliary head and a lead head, YOLOv7 achieves superior object detection accuracy \cite{RN20}. In YOLOv8, a modified CSPDarkNet53 backbone and PAN-FPN neck enable object identification, image segmentation, and movement tracking. The inclusion of Programmable Gradient Information and  Generalized Efficient Layer Aggregation Network in YOLOv9 and YOLOv10 enhances detection accuracy, with YOLOv10 also removing non-maximum suppression \cite{RN21, RN22}.

The goal of this study is to evaluate the YOLOv6 object detection model in relation to other YOLO models, focusing on accuracy and inference speed across its various variants (n, s, m, l, x). In addition to performance metrics, the research aims to analyze architectural differences, training methodologies, and computational efficiency to provide a comprehensive comparison. The study also investigates YOLOv6’s adaptability and its scalability for varying levels of complexity, offering insights into its practical usability in real-time environment.

\section{Evolution of YOLO}

\subsection{YOLO and YOLOv2}

YOLO revolutionized object detection with its speed and efficiency but faced limitations. Due to its grid-based structure, YOLO had trouble detecting smaller objects in clusters, struggled with shapes, and suffered from localization inaccuracies caused by its loss function treating small and large box errors equally \cite{RN117, RN28}. YOLOv2 employed the Darknet-19 architecture. Batch normalization enhances performance by 2\% mAP and avoids overfitting through regularization. A higher input resolution of 448×448 increases accuracy by 4\% mAP for superior performance. Anchor boxes are used instead of connected layers to enhance recall. K-means clustering optimized the precision-recall balance via anchor box selection \cite{RN29}.

\subsection{YOLOv3}

YOLOv3 introduces Darknet-53, a new 106-layer network architecture featuring residual blocks and upsampling networks. This design is larger, faster, and more accurate than the previous Darknet-19 used in YOLOv2. Key advancements include improved bounding box predictions through a logistic regression model, enabling more precise objectness scoring. For class predictions, YOLOv3 replaces the softmax approach with independent logistic classifiers, allowing better handling of complex scenarios with overlapping labels. Additionally, YOLOv3 enhances prediction accuracy by performing three-scale predictions for each location in the input image, capturing fine-grained semantic details for superior output quality \cite{RN30}.

\subsection{YOLOv4}

YOLOv4 achieves a remarkable balance of speed and accuracy in object detection, outperforming YOLOv3 with a 10\% improvement in accuracy and a 12\% boost in speed. With 29 convolutional layers and 3 × 3 filters, CSPDarknet53 has an estimated 27.6 million parameters. Key enhancements over YOLOv3 include the addition of a Spatial Pyramid Pooling block, which expands the receptive field and isolates important contextual features without slowing down the network. YOLOv4 replaces YOLOv3’s Feature Pyramid Network (FPN) with PANet to improve parameter aggregation across detection layers. It introduces advanced data augmentation techniques, such as the mosaic method. Hyper-parameters are optimized using genetic algorithms, further boosting the model’s performance and robustness \cite{RN31}.

\subsection{YOLOR}

You Only Look One Representation (YOLOR) \cite{RN118} is a unified network for multiple tasks, combining explicit (conscious) and implicit (experiential) knowledge to enhance performance. By aligning features, optimizing predictions, and standardizing representations for multi-task learning, this integration enhances the architecture. Precision improves by roughly 0.5\% in YOLOR by embedding implicit representation in each FPN’s feature map, with further refinement of predictions coming from its inclusion in the network’s output layers. For multi-task learning, YOLOR incorporates canonical representation to mitigate the performance degradation often caused by joint optimization of loss functions across tasks, achieving better performance than other models.

\subsection{YOLOX}

The backbone of YOLOX \cite{RN119} is Darknet-53, integrated into an updated YOLOv3 structure. The decoupled head in YOLOX splits the processes of detecting objects and determining their spatial coordinates. YOLOX also incorporates advanced data manipulation for training techniques, Mosaic and MixUp. It features an anchor-free design, eliminating the need for anchor-based clustering, by minimizing time and maximizing efficiency. Moreover, YOLOX introduces SimOTA, a new label assignment strategy that replaces the Intersection of Union (IoU) approach. SimOTA not only reduces training time and avoids hyperparameter issues, but also improves the model's detection mAP by 3%.

\subsection{YOLOv5}

YOLOv5 is implemented in PyTorch instead of Darknet, featuring CSPDarknet53 as its backbone. It consists of five model sizes: YOLOv5s, YOLOv5m, YOLOv5l, and YOLOv5x. A significant architectural improvement in YOLOv5 is the inclusion of the Focus layer, which consolidates multiple operations into a single layer. This change reduces both the number of layers and parameters, while also enhancing the model’s forward and backward speeds without compromising the mean average precision (mAP). YOLOv5 lost overlapping objects; with Single Object Tracking and Multiple Object Tracking (MOT), it tracked one or multiple objects with errors, while MOT with dropout tracking was more accurate \cite{RN32}. Despite YOLOv5's lower accuracy compared to ResNet and ResNeXt, it is capable of classifying multiple categories simultaneously \cite{RN34, RN127, RN129}. 

\subsection{YOLOv6}

\begin{figure}[H]
\begin{adjustwidth}{-\extralength}{0cm}
\centering
\includegraphics[width=15cm]{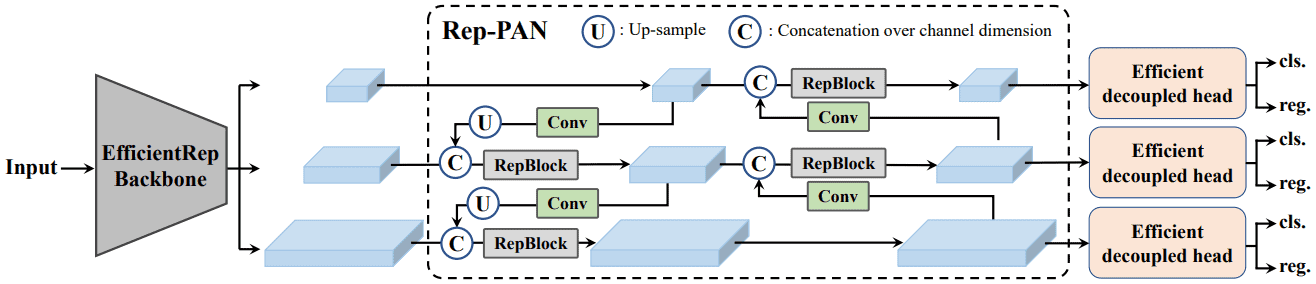}
\end{adjustwidth}
\caption{Architecture model of YOLOv6. Adapted from Rath \cite{RN36}.}
\label{Figure:1}
\end{figure}

YOLOv6 is a PyTorch-based, single-stage object detection system specifically for industrial purposes. Key improvements over YOLOv5 in this version include a hardware-optimized backbone and neck design, a better decoupled head, and a more advanced training strategy. In terms of accuracy and speed, YOLOv6 exceeds the performance of earlier YOLO models, as demonstrated by the COCO dataset. With 1234 FPS and 35.9\% AP on an NVIDIA Tesla, YOLOv6-N performed well, while YOLOv6-S set a new standard with 43.3\% AP at 869 FPS. YOLOv6-M and YOLOv6-L delivered even higher accuracy, achieving 49.5\% and 52.3\%, respectively, AP without compromising on speed (Figure \ref{Figure:1}) \cite{RN36}.

\subsection{YOLOv7}

YOLOv7 \cite{RN37} introduces significant improvements in both its architecture and its approach to trainable bag-of-freebies, setting a new trend for real-time object detection. At the architectural level, YOLOv7 integrates the Extended Efficient Layer Aggregation Network, which enables the model to learn more diverse features for better overall performance. To accommodate varying inference speed requirements, the architecture is scaled by incorporating features from YOLOv4, Scaled YOLOv4, and YOLO-R. The concept of trainable bag-of-freebies involves boosting the model's accuracy without increasing training costs, which contributes to both faster inference and improved detection accuracy in YOLOv7.

\subsection{YOLOv8}

With a more modular and adaptable architecture, YOLOv8 \cite{RN38, RN128} enables straightforward customization and fine-tuning for a wide range of applications such as object detection, segmentation, and pose estimation. YOLO-NAS (Neural Architecture Search) \cite{RN39} uses an automated approach to design better architectures, enhancing performance without the need for manual intervention. It is specifically designed to balance high accuracy with efficient resource usage, making it ideal for both accurate models and low-latency applications. Furthermore, YOLO-NAS dynamically adjusts image resolution for different objects, further optimizing inference efficiency.

\subsection{YOLOv9}

YOLOv9 \cite{RN120} introduces techniques for real-time object detection, including Programmable Gradient Information, which optimizes gradient flow during training for more efficient learning from difficult datasets. The Generalized Efficient Layer Aggregation Network improves accuracy and speed, enhancing feature extraction and aggregation. By setting new benchmarks on the MS COCO dataset, YOLOv9 outshines earlier versions in precision and its adaptability to a range of tasks. With its foundation in the Ultralytics YOLOv5 codebase, YOLOv9 to improve object detection delivers major advancements in training throughput, versatility, and precision for real-time tasks.

\subsection{YOLOv10}

By eliminating non-maximum suppression (NMS) in post-processing, YOLOv10 \cite{RN121} significantly boosts inference speed, marking a substantial advance in real-time object detection. With the introduction of dual label assignments in its NMS-free training, YOLOv10 achieves an ideal trade-off between speed, accuracy, and computational efficiency. Through improvements like lightweight classification heads, spatial-channel decoupled downsampling, and rank-guided block design, the architecture reduces both computational complexity and parameter count. Enhancing scalability and efficiency, these innovations make YOLOv10 versatile enough for deployment on both high-performance servers and edge devices with limited resources. Testing shows YOLOv10 outperforms YOLOv9 in latency and model size while delivering comparable or better detection accuracy.

\subsection{YOLOv11}

YOLOv11 enhances object detection through innovations in architecture and efficient feature processing. It integrates the C3K2 block, SPFF module, and C2PSA block, which improve accuracy and speed. The C3K2 block uses smaller convolutional kernels for better efficiency, while SPFF boosts detection across multiple scales, especially for small objects. The C2PSA block adds attention mechanisms, enhancing focus on critical regions in images. YOLOv11’s modular backbone, including Conv blocks and Bottle Necks, efficiently processes features. It also employs a multi-scale prediction head to detect objects of different sizes across three feature maps, improving detection precision  \cite{RN122}. To conclude, this paper concentrates mainly on the architecture of YOLOv6 models.

\section{Methods} 

The updated YOLOv6 framework includes components like label assignment strategies, network design, industry-oriented refinements, loss functions, and support for quantization and deployment  \cite{RN123, RN125, RN126}.

\subsection{Network design}

One-stage object detectors are typically structured with a backbone, a neck, and a head. In previous, classification improves with multibranch networks \cite{RN107}, but they compromise parallelism and increase latency; single-path models namely VGG \cite{RN106} ensure efficient inference with low memory. The feature representation capability of a network is mainly dictated by the backbone, which also significantly affects inference efficiency due to its highest cost. Backbones like RepVGG \cite{RN103}, using structural re-parameterization for better speed and accuracy, offer strong feature representation in small networks but face scaling issues. RepBlock \cite{RN103} is used for small networks; CSPStackRep Block  \cite{RN104} optimizes larger models.

YOLOv6's neck employs the PAN topology \cite{RN105} from revios models, upgraded with RepBlocks, enhancing small-scale models, or CSPStackRep Blocks, optimizing large model architecture, to develop RepPAN. By simplifying the head and implementing a hybrid-channel strategy, YOLOv6’s Efficient Decoupled Head reduces convolution layers, scales the head width, and lowers computation costs and inference latency. The box regression branch in the anchor point-based system predicts how far the anchor point is from the bounding box’s four sides.

\subsection{Label assignment}

In the training phase, label assignment ensures that labels are assigned to predefined anchors. SimOTA, a less complex version of OTA \cite{RN108}, defines positive/negative samples globally, simplifying it by reducing hyperparameters. However, in YOLOv6, SimOTA slowed training and caused instability, prompting a switch to Task Alignment Learning (TAL), which proved to improve performance and stabilize training. TAL, used with an Efficient Decoupled Head, resolves misalignment issues and enhances model efficiency.

\subsection{Loss functions}

In box regression, classification, anchor-free object detectors, and object losses are included, with VariFocal Loss (VFL) \cite{RN110} for classification and SIoU \cite{RN113}/GIoU \cite{RN114} Loss for regression tasks.

In order to optimize detectors, improving classifier performance is important. Focal Loss \cite{RN109} addresses class imbalance,  Quality Focal Loss (QFL) \cite{RN111} enhances localization quality, VFL balances positive and negative samples, and Poly Loss \cite{RN112} outperforms Cross-entropy, where VFL is chosen for YOLOv6.

Box regression loss is vital for precise bounding box localization. IoU-series losses, like GIoU and SIoU, are effective due to their alignment with evaluation metrics. In YOLOv6, YOLOv6-N and YOLOv6-NT utilized SIoU, and other models applied GIoU. Probability Loss Distribution Focal Loss (DFL) \cite{RN111} improves accuracy by modeling box locations as a probability distribution. However, due to additional layers, DFL and its variant DFLv2 \cite{RN115} add to the significant computational overhead, limiting their use in smaller models. DFL is utilized for YOLOv6-M and YOLOv6-L.

\subsection{Industry-handy improvements}

To optimize performance, self-distillation and more training epochs were used. Box regression distillation is achieved via DFL, while cosine decay balances soft and hard label information. Training duration was increased from 300 to 400 epochs for improved results.

During evaluation, a half-stride gray border is introduced to images to facilitate object detection at edges, even though it provides no meaningful context. However, inference speed decreases due to the presence of extra gray pixels, which originate from the gray border padding in Mosaic augmentation. Adjustments to the gray border area combined with direct resizing of images to the target dimensions help the model enhance or maintain performance without slowing inference.

\subsection{Quantization and deployment}

Quantization is a common practice for industrial deployment to accelerate runtime with minimal performance compromise. While post-training quantization (PTQ) employs a small calibration set for quantizing models, quantization-aware training (QAT) uses the complete training data and integrates distillation to improve performance. However, extensive usage of re-parameterization block usage poses challenges for efficiency of PTQ and complicates QAT integration due to difficulties between training and inference while aligning fake quantizers, emphasizing challenges and proposing practical solutions.

By incorporating gradient re-parameterization at each optimization step, RepOptimizer \cite{RN116} solves quantization problems in reparameterization-based models, leading to the restructuring of blocks and training for PTQ-friendly weights. PTQ performance is enhanced by partially converting quantization-sensitive operations to floating-point computations. Signal-to-noise ratio (SNR), cosine similarity, and Mean square error (MSE) are used to assess sensitivity, calculating the output feature map of a layer with and without quantization. These metrics are trained on RepOptimizer of the YOLOv6-S model and selected the top six sensitive layers to run in floating-point.

\section{Architecture of YOLOv6}

Object detection, one of the important tasks in YOLOv6 models, is the process of identifying and extracting significant features from input images. Edges, shapes, textures, and patterns are vital aspects of understanding the content of the image. Then, these features are fed into a predictive model, which processes to locate objects within the image and assign labels. The model identifies the positions by providing a bounding box.

YOLOv6 is designed with a primary focus on industrial applications, ensuring optimal performance by emphasizing both speed and accuracy. To cater to different requirements, YOLOv6 offers various model versions, including the ultra-fast YOLOv6-N (Nano), along with YOLOv6-T, YOLOv6-RepOpt, YOLOv6-S, YOLOv6-M, YOLOv6-L-ReLu, and the large-scale YOLOv6-L, providing flexibility for a variety of deployment scenarios. YOLOv6 adopts an anchor-free approach for object detection, making it 51\% faster than traditional anchor-based models. The model is structured into three main elements: the Backbone, Neck, and Head, optimizing the feature extraction, information aggregation, and final detection stages. It also uses the EfficientRep backbone, which includes RepBlock, RepConv, and CSPStackRep blocks to enhance performance. These design choices contribute to YOLOv6's efficiency and speed, setting it apart from earlier YOLO versions \cite{RN124} (Table \ref{tab:yolov6_architecture}).

\begin{table}[H]
\caption{YOLOv6 Architecture.\label{tab:yolov6_architecture}}
    \begin{adjustwidth}{-\extralength}{0cm}
        \begin{tabularx}{\fulllength}{CCCCC}
            \toprule
            \textbf{Layer} & \textbf{Filters} & \textbf{Size} & \textbf{Repeat} & \textbf{Output Size} \\
            \midrule
            Image          & -                & -             & -               & 640 × 640            \\
            Conv           & 16               & 3 × 3 / 2     & 1               & 320 × 320            \\
            Conv           & 32               & 3 × 3 / 2     & 1               & 160 × 160            \\
            RepBlock       & 32               & 1 × 1 / 1     & 1               & 160 × 160            \\
            Conv           & 64               & 3 × 3 / 2     & 1               & 80 × 80              \\
            RepBlock       & 64               & 1 × 1 / 1     & 2               & 80 × 80              \\
            Conv           & 128              & 3 × 3 / 2     & 1               & 40 × 40              \\
            RepBlock       & 128              & 1 × 1 / 1     & 2               & 40 × 40              \\
            Conv           & 256              & 3 × 3 / 2     & 1               & 20 × 20              \\
            CSPStackRep    & 256              & 1 × 1 / 1     & 1               & 20 × 20              \\
            Upsample       & -                & 2 ×           & 1               & 40 × 40              \\
            Concat         & -                & -             & 1               & 40 × 40              \\
            RepBlock       & 128              & 1 × 1 / 1     & 1               & 40 × 40              \\
            Upsample       & -                & 2 ×           & 1               & 80 × 80              \\
            Concat         & -                & -             & 1               & 80 × 80              \\
            RepBlock       & 64               & 1 × 1 / 1     & 1               & 80 × 80              \\
            Conv           & 64               & 3 × 3 / 2     & 1               & 40 × 40              \\
            Concat         & -                & -             & 1               & 40 × 40              \\
            RepBlock       & 128              & 1 × 1 / 1     & 1               & 40 × 40              \\
            Conv           & 128              & 3 × 3 / 2     & 1               & 20 × 20              \\
            Concat         & -                & -             & 1               & 20 × 20              \\
            CSPStackRep    & 256              & 1 × 1 / 1     & 1               & 20 × 20              \\
            Efficient Decoupled Head & 64, 128, 256 & -         & 1               & 80 × 80, 40 × 40, 20 × 20 \\
            \bottomrule
        \end{tabularx}
    \end{adjustwidth}
    \noindent{\footnotesize{*Note: This table summarizes the YOLOv6 architecture, featuring RepBlocks, CSPStackRep, and Efficient Decoupled Head for optimized object detection.}}
\end{table}

YOLOv6 introduces several key innovations for improved performance. It uses an anchor-free design for better generalization and reduced post-processing time. To increase efficiency, the architecture is upgraded with a reparameterized backbone and neck. It employs Varifocal Loss (VFL) for enhanced classification and Distribution Focal Loss (DFL) for better detection accuracy. Industry-centric upgrades such as increased training periods, reduced model size via quantization, and teacher-student training using knowledge distillation make YOLOv6 an optimal solution for real-time object detection, ensuring scalability and speed in a variety of industrial settings.

The backbone of YOLOv6 is crucial for feature extraction, as it handles most of the network’s computations. YOLOv6 uses reparameterized backbones to balance efficiency and accuracy. While networks like ResNets offer high classification accuracy but are slower, linear networks like VGG are faster with effective 3x3 convolutions, though less accurate. By reparameterizing, YOLOv6 enhances both speed and performance, allowing its structure to evolve during training and inference, achieving improved results.

In most object detection models, the neck is responsible for aggregating multi-scale feature maps using Path Aggregation Networks (PAN). YOLOv6 introduces a variation, known as reparameterized PAN (Rep-PAN), which enhances feature concatenation by incorporating elements from various reparameterized blocks. This innovation improves the network's ability to aggregate features from different scales more efficiently, ensuring optimal performance in detecting objects of various sizes. For head, it utilized Efficient Decoupled Head.

\subsection{Implementation}

The Optimizer employed for training is Stochastic Gradient Descent, incorporating a cosine decay-based learning rate schedule and momentum. This combination helps the training converge effectively while mitigating gradient oscillations. The Optimizer Weight Decay strategy applies Exponential Moving Average (EMA), which smooths weight updates and improves model generalization performance. For Data Augmentation, techniques such as Mosaic and Mixup are utilized. These augmentations improve the dataset variety by blending images, which enhances model robustness for object detection tasks. The Training Data is derived from the COCO 2017 training set, a benchmark dataset for large-scale object detection tasks. Similarly, the Validation Data consists of the COCO 2017 validation set, ensuring consistent evaluation during training. The training process uses 8 NVIDIA A100 GPUs, a high-performance computational setup enabling accelerated model training. Additionally, for Speed Benchmarking, the hardware consists of an NVIDIA Tesla TensorRT v7.2, ensuring optimized inference speed measurements (Table \ref{tab:yolov6_setup}).

\renewcommand{\arraystretch}{1.4}
\begin{table}[h!]
\centering
\caption{Training and hardware setup.} % Caption is now placed above the table
\begin{tabular}{|>{\raggedright}p{5cm}|p{8cm}|}
\hline
\textbf{Optimizer} & Stochastic Gradient Descent (SGD) with momentum and cosine decay learning rate \\
\hline
\textbf{Optimizer Weight Decay} & Exponential Moving Average (EMA) \\
\hline
\textbf{Data Augmentation} & Mosaic and Mixup \\
\hline
\textbf{Training Data} & COCO 2017 training set \\
\hline
\textbf{Validation Data} & COCO 2017 validation set \\
\hline
\textbf{Training Hardware} & 8 NVIDIA A100 GPUs \\
\hline
\textbf{Speed Benchmark Hardware} & NVIDIA Tesla with TensorRT v7.2 \\
\hline
\end{tabular}
\label{tab:yolov6_setup} % The label follows the caption
\end{table}
{\centering
\vspace{-1em} % Adjust spacing above the footnote
\footnotesize{From Rath \cite{RN124}.}\par}

\subsection{Key areas of YOLOv6}

The developers of YOLOv6 conducted a series of controlled experiments to assess the effects of different changes made to the model. By removing or modifying certain components, the most significant impact on the model's performance has been determined. The following elements helped identify the most effective innovations in YOLOv6's design and training: (1) network architecture, (2) label assignment strategies, (3) loss functions, (4) industry-handy improvements, and (5) quantization strategies.

\subsubsection{Network architecture}

This paper evaluates the effect of distinct backbone and neck architectures, including single-path and multi-branch, on model performance, and analyzes the influence of the channel coefficient (CC) within the CSPStackRep Block. For YOLOv6-N, the smallest model, the single-path structure demonstrates superior performance in both accuracy and speed, despite its higher computational complexity compared to the multi-branch structure. This advantage arises from its reduced memory usage and more efficient parallelism, allowing it to execute more quickly. as model size increases, the multi-branch structure begins to offer a significant performance boost in terms of both speed and accuracy. As a result, for the larger models YOLOv6-M and YOLOv6-L, the multi-branch architecture is adjusted to optimize their performance.

The specifications and performance metrics of various YOLOv6 models (N, S, M, L) are summarized in Table. Each model features different architectural blocks, such as RepBlock and CSPStackRep Block, with varied CSP Connection Coefficient (CC) configurations. YOLOv6-N achieves an APval of 35.2\% with the RepBlock and 32.7\% with the CSPStackRep Block. It reaches 1237 FPS and 1257 FPS with 4.3M and 2.3M parameters, respectively, while maintaining FLOPs of 11.1G and 5.6G. YOLOv6-S achieves an APval of 43.2\% and 43.4\%. The model achieves 499 FPS and 511 FPS, featuring parameters of 17.2M and 11.5M for RepBlock and CSPStackRep Block architectures, respectively, with 44.2G and 27.7G FLOPs. 

YOLOv6-M achieves 47.9\% APval for the RepBlock, with 137 FPS, 67.1M parameters, and 175.6G FLOPs. In comparison, with CSPStackRep Block and a CC ratio of 2/3 and 1/2, it achieves APvals of 48.1\% and 47.3\%, with 237 FPS and reduced parameters to 34.3M and 27.7M, and FLOPs to 82.2G and 68.4G. YOLOv6-L exclusively uses the CSPStackRep Block with two CC ratios (2/3 and 1/2). Both configurations attain an APval of 50.1\% but slightly differ in FPS: 149 FPS for the 2/3 ratio and 151 FPS for the 1/2 ratio. The model has parameters of 54.7M and 58.5M, respectively, with 142.7G and 144.0G FLOPs (Table \ref{tab:yolov6_models}).

\begin{table}[H]
\centering
\caption{YOLOv6 model performance and specifications.} % Caption is placed above the table
\begin{tabular}{|l|l|c|c|c|c|c|}
\hline
\textbf{Models}      & \textbf{Block}            & \textbf{CC} & $\mathbf{AP_{val}}$ & \textbf{FPS (bs=32)} & \textbf{Params} & \textbf{FLOPs} \\
\hline
YOLOv6-N & RepBlock             & -   & 35.2\% & 1237 & 4.3M  & 11.1G \\
 & CSPStackRep Block    & 1/2 & 32.7\% & 1257 & 2.3M  & 5.6G  \\
\hline
YOLOv6-S & RepBlock             & -   & 43.2\% & 499  & 17.2M & 44.2G \\
 & CSPStackRep Block    & 1/2 & 43.4\% & 511  & 11.5M & 27.7G \\
\hline
YOLOv6-M & RepBlock             & -   & 47.9\% & 137  & 67.1M & 175.6G \\
 & CSPStackRep Block    & 2/3 & 48.1\% & 237  & 34.3M & 82.2G \\
 & CSPStackRep Block    & 1/2 & 47.3\% & 237  & 27.7M & 68.4G \\
\hline
YOLOv6-L & CSPStackRep Block    & 2/3 & 50.1\% & 149  & 54.7M & 142.7G \\
 & CSPStackRep Block    & 1/2 & 50.1\% & 151  & 58.5M & 144.0G \\
\hline
\end{tabular}
\label{tab:yolov6_models}
\end{table}
{\centering
\vspace{-1em} % Adjust spacing above the footnote
\footnotesize{From Li et al. \cite{RN126}.}\par}

In Table \ref{tab:yolov6_performance}, Conv refers to the standard convolution operation used in convolutional neural networks (CNNs). RepConv refers to a "reparameterized convolution" technique, which can reduce model parameters, improve efficiency, and assist with fine-tuning. The tested activation functions of YOLOv6-N include Rectified Linear Unit (ReLU), Sigmoid Linear Unit (SiLU), and Leaky ReLU (LReLU). The AP-val (accuracy) for YOLOv6-N ranges from 34.8\% to 36.6\%. The FPS (inference speed) varies, with the highest being 1246 and the lowest being 963, showcasing a balance between computational cost and accuracy. YOLOv6-M shows significantly better performance in terms of AP-val, reaching up to 48.9\%, compared to YOLOv6-N. Its FPS ranges from 180 to 236, demonstrating a good balance between inference speed and model accuracy.

\begin{table}[H]
\centering
\caption{Performance of YOLOv6-N and YOLOv6-M models with different convolution types and activation functions.}
\label{tab:yolov6_performance}
\begin{tabular}{|l|l|l|c|c|}
\hline
\textbf{Model} & \textbf{Conv.} & \textbf{Act.} & $\mathbf{AP_{val}}$ & \textbf{FPS (bs=32)} \\
\hline
YOLOv6-N & Conv & SiLU & 36.6\% & 963 \\
 & RepConv & SiLU & 36.5\% & 971 \\
 & Conv & ReLU & 34.8\% & 1246 \\
 & RepConv & ReLU & 35.2\% & 1233 \\
 & Conv & LReLU & 35.4\% & 983 \\
 & RepConv & LReLU & 35.6\% & 975 \\
\hline
YOLOv6-M & Conv & SiLU & 48.9\% & 180 \\
 & RepConv & SiLU & 48.9\% & 180 \\
 & Conv & ReLU & 47.7\% & 235 \\
 & RepConv & ReLU & 48.1\% & 236 \\
 & Conv & LReLU & 48.0\% & 185 \\
 & RepConv & LReLU & 48.1\% & 187 \\
\hline
\end{tabular}
\end{table}
{\centering
\vspace{-1em} % Adjust spacing above the footnote
\footnotesize{From Li et al. \cite{RN126}.}\par}

\subsubsection{Label assignment}

Experiments on YOLOv6-N revealed that SimOTA and TAL are the most effective label assignment strategies. TAL outperformed SimOTA by 0.5\% in AP and was chosen for its stability and better accuracy. The warm-up strategy using ATSS showed similar performance, even when replaced with SimOTA.

Adaptive Training Sample Selection (ATSS) is a method, achieving an AP(val) of 32.5\%. SimOTA (Simplified OTA) optimizes target assignment strategies, improving the AP(val) to 34.5\%. Task alignment learning demonstrates the best AP(val) of 35.0\% and focuses on aligning task objectives with training to maximize performance. DW attains an AP(val) score of 33.4\%, improving performance moderately. ObjectBox achieves the least accuracy of 30.1\% (Table \ref{tab:method_ap_results}).

\begin{table}[H]
\centering
\caption{Performance of different methods on AP(val).}
\begin{tabular}{|l|c|}
\hline
\textbf{Method}                 & $\mathbf{AP_{val}}$ (\%) \\
\hline
Adaptive Training Sample Selection                            & 32.5 \\
Simplified Optimal transport assignment                          & 34.5 \\
Task alignment learning         & 35.0 \\
DW                         & 33.4 \\
ObjectBox                       & 30.1 \\
\hline
\end{tabular}
\label{tab:method_ap_results}
\end{table}
{\centering
\vspace{-1em} % Adjust spacing above the footnote
\footnotesize{From Li et al. \cite{RN126}.}\par}

\subsubsection{Loss fuctions}

YOLOv6 uses two loss functions: Varifocal Loss (VFL) for classification, which handles hard and easy examples differently, and Distribution Focal Loss (DFL) combined with SIoU or GIoU for box regression. VFL balances learning signals from positive and negative samples. DFL, effective in cases with blurred ground-truth boundaries, treats box locations as a discretized probability distribution. YOLOv6 models employ DFL for improved box localization, although experiments with DFLv2 showed no performance gains despite added computation. In Quality Focal Loss (QFL), the extended framework merges both classification scores and the quality of localization, improving supervision for object detection tasks.

YOLOv6-S achieves significantly better AP-val values compared to YOLOv6-N. The highest AP-val for YOLOv6-N is 35.4\% (with QFL), while YOLOv6-S starts at 41.5\% and goes as high as 43.2\% (with VFL). YOLOv6-M demonstrates the best performance, with an AP-val reaching 48.1\% with VFL. VFL consistently provides the highest AP-val values, followed by QFL, Focal Loss, and Poly Loss (Table \ref{tab:yolov6_classification_loss}).

\begin{table}[H]
\centering
\caption{AP-val performance of different YOLOv6 models using various classification loss functions.}
\label{tab:yolov6_classification_loss}
\begin{tabular}{|l|l|c|}
\hline
\textbf{Model} & \textbf{Classification Loss} & \textbf{AP-val} \\
\hline
\textbf{YOLOv6-N} & Focal Loss & 35.0\% \\
& Poly Loss & 34.0\% \\
& QFL & 35.4\% \\
& VFL & 35.2\% \\
\hline
\textbf{YOLOv6-S} & Focal Loss & 42.9\% \\
& Poly Loss & 41.5\% \\
& QFL & 43.1\% \\
& VFL & 43.2\% \\
\hline
\textbf{YOLOv6-M} & Focal Loss & 48.0\% \\
& Poly Loss & 46.9\% \\
& QFL & 48.0\% \\
& VFL & 48.1\% \\
\hline
\end{tabular}
\end{table}
{\centering
\vspace{-1em} % Adjust spacing above the footnote
\footnotesize{From Li et al. \cite{RN126}.}\par}

In Table \ref{tab:yolov6_loss_function}, there is a clear trend where the use of Dynamic Focal Loss (DFL) or its variant (DFLv2) results in marginal increases in AP-val across all models compared to when no loss function is used (w/o). However, the FPS tends to decrease as DFL/DFLv2 are applied, indicating a trade-off between accuracy (AP-val) and speed (FPS). YOLOv6-M demonstrates the highest AP-val values across the models, although FPS drops significantly compared to YOLOv6-S and YOLOv6-N.

\begin{table}[H]
\centering
\caption{Performance of YOLOv6 models with different loss functions.}
\label{tab:yolov6_loss_function}
\begin{tabular}{|l|l|c|c|}
\hline
\textbf{Method} & \textbf{Loss} & \textbf{AP-val} & \textbf{FPS (bs=32)} \\
\hline
YOLOv6-N  & w/o & 35.0\% & 1226 \\
 & DFL & 35.2\% & 1022 \\
       & DFLv2 & 35.2\% & 819 \\
\hline
YOLOv6-S & w/o & 42.9\% & 486 \\
 & DFL & 43.0\% & 461 \\
       & DFLv2 & 43.0\% & 422 \\
\hline
YOLOv6-M & w/o & 48.0\% & 233 \\
 & DFL & 48.2\% & 236 \\
       & DFLv2 & 48.3\% & 226 \\
\hline
\end{tabular}
\end{table}
{\centering
\vspace{-1em} % Adjust spacing above the footnote
\footnotesize{From Li et al. \cite{RN126}.}\par}

\subsubsection{Industry-handy improvements}

YOLOv6-N shows a 35.9\% AP-val at 300 epochs and increases to 36.3\% at 400 epochs, reflecting a slight improvement as training progresses. YOLOv6-T shows significant improvement over YOLOv6-N. With 300 epochs, it achieves an AP-val of 40.3\%, and with 400 epochs, the performance rises to 40.9\%. YOLOv6-S shows the highest performance among the three. At 300 epochs, it achieves 43.4\%, and at 400 epochs, it improves to 43.9\%. Training for more epochs boosts accuracy, increasing AP by 0.4\%, 0.6\%, and 0.5\% for YOLOv6-N, T, and S, respectively, with 400 epochs offering optimal convergence (Table \ref{tab:yolov6_300_400_epochs}).

\begin{table}[H]
\centering
\caption{Performance comparison of YOLOv6 models for 300 and 400 epochs.}
\label{tab:yolov6_300_400_epochs}
\begin{tabular}{|l|c|c|}
\hline
\textbf{Model} & \textbf{300 Epochs} & \textbf{400 Epochs} \\
\hline
YOLOv6-N & 35.9\% & 36.3\% \\
YOLOv6-T & 40.3\% & 40.9\% \\
YOLOv6-S & 43.4\% & 43.9\% \\
\hline
\end{tabular}
\end{table}
{\centering
\vspace{-1em} % Adjust spacing above the footnote
\footnotesize{From Li et al. \cite{RN126}.}\par}

\subsubsection{Quantization results}

Using YOLOv6-S to validate quantization method, the baseline model for 300 epochs is trained. RepOptimizer improves performance, while partial quantization aware training (QAT) with fake quantizers on non-sensitive layers enhances accuracy and speed through graph optimization. The RepOptimizer clearly improves the model’s PTQ INT8 performance, especially in version 2.0, showing that the optimizations make a significant difference in both computational efficiency and accuracy. The model achieves higher performance across both FP32 and PTQ modes, demonstrating the advantage of the optimizations implemented (Table \ref{tab:ptq_evaluation}).

\begin{table}[H]
\centering
\caption{PTQ evaluation of YOLOv6s trained using RepOptimizer.}
\label{tab:ptq_evaluation}
\begin{tabular}{|l|c|c|}
\hline
\textbf{Model} & \textbf{FP32 AP-val} & \textbf{PTQ INT8 AP-val} \\
\hline
YOLOv6-S (v1.0) & 42.4 & 35.0 \\
YOLOv6-S w/ RepOptimizer (v1.0) & 42.4 & 40.9 \\
YOLOv6-S (v2.0) & 43.4 & 41.3 \\
YOLOv6-S w/ RepOptimizer (v2.0) & 43.1 & 42.6 \\
\hline
\end{tabular}
\end{table}
{\centering
\vspace{-1em} % Adjust spacing above the footnote
\footnotesize{From Li et al. \cite{RN126}.}\par}

YOLOv6-S model, when evaluated using INT8 quantization, achieves an AP-val of 35.0\% and a FPS of 556. YOLOv6-S + RepOpt + Partial QAT + CW Distill incorporates the following multiple optimization strategies: RepOptimizer (RepOpt), Partial Quantization-Aware Training (QAT), and Class-wise Distillation (CW Distill). With these techniques, the model improves its AP-val to 42.3\%, though FPS drops to 503. YOLOv6-S + RepOpt + QAT + CW Distill comprises complete QAT (Quantization-Aware Training) and achieves an AP-val of 42.1\%, a small decrease compared to the previous version, but it achieves a slightly higher FPS of 528 (Table \ref{tab:optimization}).

\begin{table}[H]
\centering
\caption{Comparison of YOLOv6-S performance with different optimization techniques.}
\label{tab:optimization}
\begin{tabular}{|l|c|c|}
\hline
\textbf{Model} & \textbf{INT8 AP-val} & \textbf{FPS} \\
\hline
YOLOv6-S & 35.0 & 556 \\
YOLOv6-S + RepOpt + Partial QAT + CW Distill & 42.3 & 503 \\
YOLOv6-S + RepOpt + QAT + CW Distill & 42.1 & 528 \\
\hline
\end{tabular}
\end{table}
{\centering
\vspace{-1em} % Adjust spacing above the footnote
\footnotesize{From Li et al. \cite{RN126}.}\par}

YOLOv5-S shows an AP-val of 36.9\%, with a performance of 502 FPS when batch size is 1. The N/A in the FPS (bs=32) column indicates that these results were not available for a batch size of 32. YOLOv6-S achieves an AP-val of 41.3\% with 579 FPS at batch size 1. FPS for a batch size of 32 is unavailable. YOLOv7-Tiny has a slightly lower AP-val of 37.0\% but shows 512 FPS with batch size 1, with no data for batch size 32. YOLOv6-S (FP16) shows an impressive 43.4\% AP-val, with 377 FPS for batch size 1 and 541 FPS for batch size 32. YOLOv6-S (Our QAT strategy) achieves an AP-val of 43.3\% (almost equal to the FP16 model) but outperforms in FPS, with 596 FPS at batch size 1 and 869 FPS at batch size 32 (Table \ref{tab:qat_performance}).

\begin{table}[H]
\centering
\caption{QAT Performance of YOLOv6-S (v2.0) compared to other object detectors.}
\label{tab:qat_performance}
\begin{tabular}{|l|c|c|c|}
\hline
\textbf{Model} & \textbf{AP-val} & \textbf{FPS (bs=1)} & \textbf{FPS (bs=32)} \\
\hline
YOLOv5-S & 36.9 & 502† & N/A \\
YOLOv6-S* & 41.3 & 579† & N/A \\
YOLOv7-Tiny & 37.0 & 512† & N/A \\
YOLOv6-S (FP16) & 43.4 & 377† & 541† \\
YOLOv6-S (Our QAT strategy) & 43.3 & 596† & 869 \\
\hline
\end{tabular}
\end{table}
{\centering
\vspace{-1em} % Adjust spacing above the footnote
\footnotesize{The ‘*’ refers to the results of YOLOv6-S based on the v1.0 release. The ‘†’ indicates that the performance was tested using TensorRT 8 on a Tesla T4 GPU with batch size 1 and 32. From Li et al. \cite{RN126}.}\par}

\subsection{Performance metrics}

The YOLOv5 series models (YOLOv5-N, YOLOv5-S, YOLOv5-M, YOLOv5-L) show increasing AP-val and FPS performance with model size, with YOLOv5-L being the most accurate, yielding a 67.3\% AP-val at the cost of relatively lower FPS (126 with a batch size of 32). These models are designed for scalability and show progressive improvements in performance as the model complexity increases.

 The larger input sizes (like 1280x1280) in YOLOv5-N6 through YOLOv5-L6 produce higher accuracy levels, but these models also incur an increase in the number of parameters and FLOPs. For instance, YOLOv5-X6, the largest in the series, achieves 55.0\% AP-val with 17 FPS, but with an exceptionally high 839.2 G FLOPs.

YOLOX models (YOLOX-Tiny, YOLOX-S, YOLOX-M, and YOLOX-L) exhibit comparable performance, with the highest accuracy achieved by YOLOX-L, which reaches a 68.0\% AP-val but at a cost of increased computational complexity, as reflected by the parameters and FLOPs.

PPYOLOE models present competitive performance, with PPYOLOE-L reaching 68.6\% AP-val and a lower FPS than some of the newer YOLOv6 models, but still a best choice in terms of accuracy and latency. YOLOv7 models also follow similar trends to YOLOv6, where larger variants like YOLOv7 improve accuracy but with higher latency and computational costs.

YOLOv8-N (the smallest model in the series) achieves an AP-val of 37.3\%, along with a high FPS of 561 when the batch size is 1, making it a good option for applications that need speed. It has relatively low computational costs, with only 8.7 G FLOPs, which contributes to its high FPS. YOLOv8-S increases in accuracy to 44.9\% AP-val, while its FPS drops to 311 at batch size 1. YOLOv8-M sees an increase in accuracy, reaching 50.2\% AP-val, with FPS dropping to 143 at batch size 1. YOLOv8-L offers the highest accuracy at 52.9\% AP-val, achieving a good balance between accuracy and inference speed. It achieves 91 FPS at batch size 1 but also incurs a considerable increase in computational overhead, with 165.2 G FLOPs and a latency of 11.0 ms (Table \ref{tab:yolo_model_comparison}).

\begin{table}[H]
\caption{Performance comparison of YOLO models.\label{tab:yolo_model_comparison}}
\setlength{\tabcolsep}{3pt} % Adjust column spacing
\renewcommand{\arraystretch}{1.1} % Adjust row spacing
\footnotesize % Slightly reduce font size
\begin{adjustwidth}{-\extralength}{0cm}
    \begin{tabularx}{\fulllength}{>{\raggedright\arraybackslash}p{3cm} 
                                     >{\centering\arraybackslash}p{1.5cm}
                                     *8{>{\centering\arraybackslash}X}}
        \toprule
        \textbf{Method} & \textbf{Input Size} & \textbf{$AP^\text{val}$} & \textbf{$AP^\text{val}_{50}$} & \textbf{FPS (bs=1)} & \textbf{FPS (bs=32)} & \textbf{Latency (bs=1)} & \textbf{Params} & \textbf{FLOPs} \\
        \midrule
        \textbf{YOLOv6-N}   & 640 & 37.0\% / 37.5\%‡  & 52.7\% / 53.1\%‡  & 779  & 1187 & 1.3 ms  & 4.7 M  & 11.4 G \\
        \textbf{YOLOv6-S}   & 640 & 44.3\% / 45.0\%‡  & 61.2\% / 61.8\%‡  & 339  & 484  & 2.9 ms  & 18.5 M & 45.3 G \\
        \textbf{YOLOv6-M}   & 640 & 49.1\% / 50.0\%‡  & 66.1\% / 66.9\%‡  & 175  & 226  & 5.7 ms  & 34.9 M & 85.8 G \\
        \textbf{YOLOv6-L}   & 640 & 51.8\% / 52.8\%‡  & 69.2\% / 70.3\%‡  & 98   & 116  & 10.3 ms & 59.6 M & 150.7 G \\
        \textbf{YOLOv6-N6}  & 1280 & 44.9\% & 61.5\%  & 228  & 281  & 4.4 ms  & 10.4 M & 49.8 G \\
        \textbf{YOLOv6-S6}  & 1280 & 50.3\% & 67.7\%  & 98   & 108  & 10.2 ms & 41.4 M & 198.0 G \\
        \textbf{YOLOv6-M6}  & 1280 & 55.2\%‡ & 72.4\%‡  & 47   & 55   & 21.0 ms & 79.6 M & 379.5 G \\
        \textbf{YOLOv6-L6}  & 1280 & 57.2\%‡ & 74.5\%‡  & 26   & 29   & 38.5 ms & 140.4 M & 673.4 G \\
        \midrule
        YOLOv5-N   & 640 & 28.0\%  & 45.7\%  & 602  & 735   & 1.7 ms  & 1.9 M  & 4.5 G \\
        YOLOv5-S   & 640 & 37.4\%  & 56.8\%  & 376  & 444   & 2.7 ms  & 7.2 M  & 16.5 G \\
        YOLOv5-M   & 640 & 45.4\%  & 64.1\%  & 182  & 209   & 5.5 ms  & 21.2 M & 49.0 G \\
        YOLOv5-L   & 640 & 49.0\%  & 67.3\%  & 113  & 126   & 8.8 ms  & 46.5 M & 109.1 G \\
        YOLOv5-N6  & 1280 & 36.0\%  & 54.4\%  & 172  & 175   & 5.8 ms  & 3.2 M  & 18.4 G \\
        YOLOv5-S6  & 1280 & 44.8\%  & 63.7\%  & 103  & 103   & 9.7 ms  & 12.6 M & 67.2 G \\
        YOLOv5-M6  & 1280 & 51.3\%  & 69.3\%  & 49   & 48    & 20.1 ms & 35.7 M & 200.0 G \\
        YOLOv5-L6  & 1280 & 53.7\%  & 71.3\%  & 32   & 30    & 31.3 ms & 76.8 M & 445.6 G \\
        YOLOv5-X6  & 1280 & 55.0\%  & 72.7\%  & 17   & 17    & 58.6 ms & 140.7 M & 839.2 G \\
        \midrule
        YOLOX-Tiny  & 416 & 32.8\%  & 50.3\%* & 717  & 1143  & 1.4 ms  & 5.1 M  & 6.5 G \\
        YOLOX-S     & 640 & 40.5\%  & 59.3\%* & 333  & 396   & 3.0 ms  & 9.0 M  & 26.8 G \\
        YOLOX-M    & 640 & 46.9\%  & 65.6\%* & 155  & 179   & 6.4 ms  & 25.3 M & 73.8 G \\
        YOLOX-L    & 640 & 49.7\%  & 68.0\%* & 94   & 103   & 10.6 ms & 54.2 M & 155.6 G \\
        \midrule
        PPYOLOE-S  & 640 & 43.1\%  & 59.6\%  & 327  & 419   & 3.1 ms  & 7.9 M  & 17.4 G \\
        PPYOLOE-M  & 640 & 49.0\%  & 65.9\%  & 152  & 189   & 6.6 ms  & 23.4 M & 49.9 G \\
        PPYOLOE-L  & 640 & 51.4\%  & 68.6\%  & 101  & 127   & 10.1 ms & 52.2 M & 110.1 G \\
        \midrule
        YOLOv7-Tiny & 416 & 33.3\%* & 49.9\%* & 787  & 1196  & 1.3 ms  & 6.2 M  & 5.8 G \\
        YOLOv7-Tiny & 640 & 37.4\%* & 55.2\%* & 424  & 519   & 2.4 ms  & 6.2 M  & 13.7 G* \\
        YOLOv7     & 640 & 51.2\%  & 69.7\%*  & 110  & 122   & 9.0 ms  & 36.9 M & 104.7 G \\
        \midrule
        YOLOv8-N     & 640 & 37.3\%  & 52.6\%* & 561  & 734   & 1.8 ms  & 3.2 M  & 8.7 G \\
        YOLOv8-S     & 640 & 44.9\%  & 61.8\%* & 311  & 387   & 3.2 ms  & 11.2 M & 28.6 G \\
        YOLOv8-M     & 640 & 50.2\%  & 67.2\%* & 143  & 176   & 7.0 ms  & 25.9 M & 78.9 G \\
        YOLOv8-L     & 640 & 52.9\%  & 69.8\%* & 91   & 105   & 11.0 ms & 43.7 M & 165.2 G \\
        \bottomrule
    \end{tabularx}
\end{adjustwidth}
\end{table}
{\centering
\vspace{-1em} % Adjust spacing above the footnote
\footnotesize{‡ AP values at higher IOU thresholds. * Denotes results based on the batch size 1 or other test variations. From Li et al. \cite{RN125}.}\par}

YOLOv6 series models (YOLOv6-N, YOLOv6-T, YOLOv6-S, YOLOv6-M, and YOLOv6-L) perform well overall, especially in terms of latency and FPS. YOLOv6-N achieves a 51.2\% AP-val with 1234 FPS on batch size 32, demonstrating both good accuracy and processing speed. YOLOv6-L shows the highest AP-val of 70.0\% and has a relatively good FPS of 121. Despite the improved accuracy, it comes with an increased latency of 10.2ms and higher computational costs (58.5 million parameters and 144.0 billion FLOPs).

The inclusion of YOLOv6-N6 through YOLOv6-L6, which utilize larger input sizes (1280x1280), further scales up the accuracy, albeit with an increase in parameters and FLOPs. For instance, YOLOv6-L6 offers a significant jump in performance to 57.2\% AP-val but also demands higher processing time (38.5 ms latency) and considerably more computational power (673.4 G FLOPs).

Overall, the YOLOv6 series is great at balancing accuracy and speed, making it the best choice for real-time detection tasks, especially when fast performance is important. YOLOv6 models maintain a strong FPS performance, reduce latency compared to competing models like YOLOv5, and scale efficiently across a range of deployment needs. Hence, YOLOv6 proves to be highly suitable for a wide array of real-time detection scenarios, specifically industrial environments, without significant compromises on either processing speed or accuracy.

\section{Conclusions}

In conclusion, with a primary focus on industrial demands, YOLOv6 advances in object detection to deliver efficient real-time detection capabilities. With superior network design and refined training strategies, the model achieves better accuracy and speed than competing real-time detectors. The inclusion of a custom quantization method in YOLOv6 boosts its efficiency, making it well-suited for industrial environments. It is anticipated that over time the use of lightweight object detectors will have a transformative impact on automating domains such as renewable energy ~\cite{hussain2019deployment}, emotional intelligence ~\cite{aydin2023domain,hussain2023child}, security and industrial quality inspection.

%%%%%%%%%%%%%%%%%%%%%%%%%%%%%%%%%%%%%%%%%%
\begin{adjustwidth}{-\extralength}{0cm}
%\printendnotes[custom] % Un-comment to print a list of endnotes

\bibliographystyle{unsrt}  % Changes bibliography style to unsorted
\bibliography{ref}  % This points to the filename of your BibTeX file without the .bib extension

\end{adjustwidth}
\end{document}